\documentclass[pdflatex,sn-mathphys]{sn-jnl}

\jyear{2021}%

\theoremstyle{thmstyleone}%
\theoremstyle{thmstyletwo}%

\newcommand{\comment}[1]{}

\theoremstyle{thmstylethree}%

\newenvironment{itemizerCompact}{\vspace{-1mm}
  \begin{itemize}
    \setlength{\itemsep}{2pt}
    \setlength{\parskip}{0pt}
    \setlength{\parsep}{0pt}
  }
{ \end{itemize}
  \vspace{-1mm}  }

\begin{document}

\title[An overview of AI in CALL]{An overview of artificial intelligence
  in computer-assisted language learning}


\author{\fnm{Anisia} \sur{Katinskaia}}\email{first.last@helsinki.fi}

\affil{\orgname{University of Helsinki}, \orgaddress{\country{Finland}}}
\affil{\orgdiv{Department of Computer Science}}
\affil{\orgdiv{Department of Digital Humanities}}


\abstract{Computer-assisted language learning---CALL---is an established research field. 
We review how {\em artificial intelligence} can be applied to support language learning and teaching. 
The need for intelligent agents that assist language learners and teachers is increasing: the human teacher's time is a scarce and costly resource, which does not scale with growing demand.  Further factors contribute to the need for CALL: pandemics and increasing demand for distance learning, migration of large populations, the need for sustainable and affordable support for learning, etc.

CALL systems are made up of many components that perform various functions, and AI is applied to many different aspects in CALL, corresponding to their own expansive research areas.
Most of what we find in the research literature and in practical use are prototypes or partial implementations---systems that perform some aspects of the overall desired functionality.
{\em Complete} solutions---most of them commercial---are few, because they require massive resources.  

Recent advances in AI should result in improvements in CALL, yet there is a lack of surveys that focus on AI in the context of this research field.
This paper aims to present a perspective on the AI methods that can be employed for language learning from a position of a developer of a CALL system.  We also aim to connect work from different disciplines, to build bridges for interdisciplinary work.}

\keywords{CALL, AI, language learning, Intelligent Tutoring Systems, ITS}



\maketitle

\section{Introduction}
\label{sec1}

Intelligent assistance for second language (L2)---or foreign language---learning and teaching poses a significant research challenge.  Computer-aided language learning (CALL) is a field integrating research from various disciplines, including artificial intelligence and machine learning, language technology, educational data science (EDS), applied linguistics, and language pedagogy~\cite{meurers2012natural}.  Design and software engineering also play an important role: an engaging user experience (UX) is vital for maintaining learner motivation and encouraging continued use of the CALL system.

The concept of employing computers for language education dates back to 50 years ago.  As an active area of research, CALL evolved from the broader domain of computer-assisted instruction.  CALL is broadly defined as ``the search for and study of applications of the computer in language teaching and learning,''~\cite{levy1997computer}.  Among the pioneering CALL systems was \textit{PLATO}~\cite{chapelle1983language,hart1981language},  developed in the early 1970s, which stood out as one of the earliest and most significant tools for teaching and learning languages.  \comment{\textit{Macario} was one of the first video-based programs for learning Spanish~\cite{gale1989macario}; the \textit{Athena Language-Learning Project} (ALLP) combined interactive elements with more traditional
drill-and-practice routines~\cite{morgenstern1986athena}; and programs like
\textit{\`A la rencontre de Phillippe}~\cite{murray2014knowledge} provided an environment for engaging with the target language.}

Within the CALL domain, the ``computer-as-a-tutor'' modality has been widely acknowledged and has been a focal point for decades, despite the prominence of the ``computer-as-a-tool'' modality~\cite{slavuj2015intelligent,tafazoli2019intelligent}. \comment{heift2003student} Our discussion will center on the tutor modality, with a focus on {\em structure and functionality} of intelligent tutoring systems (ITS) in the context of language learning.  ITS originates from the development of Intelligent Computer-Assisted Learning (ICAL) in the 1960s and 1970s~\cite{anderson1985intelligent,graesser2012intelligent}.  It has recently been deployed across various learning domains, including high-school and college-level mathematics, sciences, business management, etc.   ITS is characterized by its capacity to generalize and apply knowledge to specific tasks and to adapt dynamically based on the user and the tasks at hand.  Initially, the goal of ITS was to emulate the role of human tutors---not only to facilitate learning but also to monitor and respond to the learner's progress~\cite{self1990theoretical,self1998defining}.  \comment{shute1991rose} 

Decades of research have shown the effectiveness of ITS technology in enhancing learning.  Studies indicate that students who engage with ITSs often demonstrate greater performance gains than those limited to traditional classroom environments~\cite{kulik2016effectiveness}.  Prominent systems like \textit{Cognitive Tutor}~\cite{ostrow2014testing,oxman2014white}, \textit{ALEKS}~\cite{craig2013impact},
\textit{ASSISTments}~\cite{koedinger2010quasi} were used by tens or hundreds of thousands of students annually, significantly elevating learning outcomes.  Among the ITS developed for language learning, we can name systems such as \textit{E-Tutor}, for learning university-level German~\cite{heift2008modeling,heift2010developing}; \textit{TAGARELA}, for learning university-level Portuguese~\cite{amaral2007conceptualizing,amaral2011using}; and \textit{Robo-Sensei}, for learning Japanese, mostly focusing on translation~\cite{nagata2009robo}.  The first two systems employ adaptive learner models to provide tailored feedback messages, depending on individual performance.  The third system offers a fixed sequence of activities across 24 lessons, without adaptation to individual learner profiles.  These systems are monolingual and, as of the writing of this thesis, are not accessible for use.

The advancement of technology, especially in machine learning, has boosted the application of ITS across various educational domains, including language learning.  Ideally, an ITS would provide a personalized learning experience, mirroring a real teacher who\comment{knows everything about the domain and} can tailor and steer the learning process to the individual needs of each student for the mastery of specific skills.  However, given the complexity of learning a language, replacing a human teacher with ITS is unrealistic.  Thus, even if the ITS could undertake a modest portion of the teacher's tasks---10-15\%---it would already result in substantial savings of human resources.

The {\em structure} of an ITS comprises three core components: the {\em Domain Model}, the {\em Student Model}, and the {\em Instruction Model} (or Tutoring Model)~\cite{Polson1988foundations,wenger2014artificial}.  With these models defined, the ITS is considered to be fully specified. A four-model architecture retains the above-mentioned components and adds the {\em User Interface Model}~\cite{padayachee2002intelligent}.  ITS {\em functions} that are considered to be important for the success of CALL include: (1) {\em generation of exercises} that target both productive and receptive skills across varying proficiency levels; (2) detailed {\em assessment} of the learner's skills/performance, the learning material's difficulty, and individual learning items; (3) provision of {\em feedback} that the ITS communicates to the learner throughout the learning process; (4) and the {\em analytics} that inform the learner and teacher during the learning process, potentially outlining a learning pathway toward the desired proficiency levels.  The level of proficiency of the target learner group is pivotal during CALL system development, as strategies effective for beginners may substantially differ from those appropriate at intermediate and advanced levels~\cite{ebadi2017developing,felix2007pragmatic, van2005instructional}.

The interplay between the ITS components and functions is extensive.  Inter-relationships among them can build bridges for interdisciplinary work, which will be, we believe, a central focus in future CALL research.

The paper is organized as follows.  In Section~\ref{structure}, we describe the tutoring system structure in more detail.  In Section~\ref{functions}, we cover tutoring system functions that can be seen as components of the Instruction model, and which connect all three ITS models together.  Section~\ref{conclusions} presents conclusions and future work.

\section{System structure}
\label{structure}

ITS literature states that to simulate a good teacher, the ITS needs to know:
\begin{itemizerCompact}
\item the subject matter,
\item the learner's current state, and
\item the tasks or exercises most suitable for the learner---that will lead toward improvements in proficiency most effectively~\cite{albacete2019impact,graesser2017assessment,mu2018combining}. \comment{munshi2019personalization}
\end{itemizerCompact}

These correspond to the Domain, Student, and Instruction Models in the ITS. It also should be able to interact with the learner (Interface Model) to present generated tasks, feedback, and results of the learner's assessment in the clearest and most constructive way, as well as to keep the learner motivated and engaged.

\paragraph{Domain Model} The {\em Domain Model} represents the knowledge structure of the learning domain within an intelligent tutoring system.  The Domain model typically encompasses various types of knowledge, such as declarative and procedural knowledge~\cite{anderson:cognitive}. \comment{Declarative knowledge includes facts, concepts, and principles, whereas procedural knowledge comprises sets of production rules and algorithms that define the skill.} It is a representation of everything that must be mastered by the learner: facts, rules, concepts, etc., (known as {\em skills} in ITS literature)---and the {\em relationships} among them\comment{, e.g., prerequisite relation which is known as the ``surmise'' relation in ITS}~\cite{Polson1988foundations,wenger2014artificial}.  For language learning, the Domain model includes linguistic phenomena or rules ranging in specificity.  For example, synonyms of the word {\em``retain''}, the rule (in English) ``uncountable nouns usually take no indefinite article'' (? {\em ``a sand''}), word derivation (forming words from simpler words), inflection of irregular verbs, syntactic relations between verbs and their governees, various collocations, idioms, etc.  
Within this thesis, we will use the term {\em construct} to refer to all of these items that constitute the Domain model in language learning.

The representation of domain knowledge can vary across learning domains, adopting formats like semantic networks or ontologies~\cite{chang2020building,kazi2010leveraging,stankov2000knowledge}.\comment{ahmed2020ontology} As mentioned by Slavuj et al.~\cite{slavuj2015intelligent}, the most common method for representing domain knowledge in ITS for language learning is a network of items, linked by various relations.  Typically, this knowledge originates from human experts: textbooks, grammar, etc.  However, the Domain model can also be refined using knowledge extracted from extensive learner data, in particular by looking for patterns of correct and incorrect answers given by learners, to infer information about potential dependencies and learning trajectories.  Pattern recognition from learner data is an area of focus within educational data mining~\cite{aldowah2019educational,romero2020educational}.
A comprehensive review of educational data mining for L2 learning can be found in the work of Bravo et al.~\cite{bravo2020data}.

\paragraph{Student Model} The {\em Student model} represents the inferred ``state'' of the learner’s knowledge and skill at any given time in the learning process.  Static characteristics of the learner include mother tongue, previously studied languages, etc.  Dynamic characteristics include information about learner proficiency: performance on tasks, errors, timing (time required to answer questions, exercises, etc.), preferences about content, learning styles \cite{10.5120/ijca2019918451,slavuj2015intelligent}.  This part of the ITS captures all features that distinguish one learner from another and enables the ITS to {\em personalize} the learning experience.  The Student Model utilizes the Domain Model to represent which parts of the domain the learner has mastered and to what extent, as well as what requires further practice.

\paragraph{Instruction Model} The {\em Instruction Model}, or {\em Tutor Model}, serves as the {\em pedagogical} component, determining what to teach, when to teach it, and how to teach it---based on the information from the Student and Domain models.  This model orchestrates the selection and sequencing of constructs and tasks to be introduced to the learner next. 
This relates to the notion of the ``Zone of Proximal Development'' (ZPD) in education science---the range of tasks that the student is ready to learn next with appropriate assistance~\cite{shabani2010vygotsky,vygotsky:1978:proximal}. \comment{In Vygotsky's socio-cultural theory, identifying the ZPD at every step in the learning process is essential for effective learning.} For a given learner, some constructs may be overly challenging, while others might be insufficiently stimulating.  Adhering to the ZPD principle means focusing on tasks that strike a balance---neither too simple, which will bore the learner, nor too difficult, which will cause frustration.  Therefore, the system must predict which constructs are optimally suited for the learner's current proficiency level, which involves continuous {\em assessment} of performance. The Instruction model interacts dynamically with the other components: it generates new exercises based on the Domain and Student Models, offers feedback in response to the learner's actions, and refines the Student Model based on ongoing performance data.

\section{System functions}
\label{functions}

In this section, we cover the automatic generation of grammar exercises, along with exercises for reading, listening, and speaking. The second part focuses on automatic assessment and feedback generation, particularly in the areas of automatic essay scoring and speech evaluation.

\subsection{Exercise generation}

\comment{The component of the learning system responsible for {\em exercise generation} relates to both the Domain and the Student model.  Given that creating questions and exercises manually is a time-consuming and tedious task, there is a significant interest in the automatic generation of questions across educational disciplines.}  

The forms of automatically generated exercises for language learning prevalent in literature are wh-questions (the answers are short facts directly stated in the input sentence), gap-filling questions (fill-in-the-blank) based on text, and multiple-choice questions~\cite{kurdi2020systematic}.  Gap-filling exercises can be designed as a gap with or without a hint, posing varying levels of challenge~\cite{perez-beltrachini-etal-2012-generating}.  Multiple-choice questions can be seen as a gap-filling exercise with available options to choose from.  In language learning, the objectives of these exercises can be formulated as, but are not limited to, the following:
\begin{itemize}
    \item challenge the reader’s comprehension skills~\cite{hill2016automatic},
    \item learning vocabulary~\cite{jiang2017distractor,qiu2021automatic}, and
    \item learning grammar, e.g., the use of prepositions~\cite{lee2016call}.
\end{itemize}

Several studies have concentrated on the task of \textbf{selecting words as targets of gap-filling exercises}.  Marrese-Taylor et al.~\cite{marrese-taylor-etal-2018-learning} formalize the problem of selecting words for fill-in-the-blank exercises as sequence labeling and sequence classification tasks, employing LSTM~\cite{hochreiter1997long} models trained on anonymized learner data from a language learning platform.  BERT~\cite{devlin-etal-2019-bert} and the TextRank~\cite{mihalcea-tarau-2004-textrank} algorithm have been applied to identify the most relevant sentences and the most important keywords for gap-filling exercises aimed at enhancing reading comprehension~\cite{yang2021automatic}.  

Currently, a range of Transformer models are used to generate open cloze exercises~\cite{felice-etal-2022-constructing,matsumori2022mask}.  For instance, Felice et al.~\cite{felice-etal-2022-constructing} propose a Transformer model for choosing tokens for gap-filling exercises: the model is trained to classify tokens as ``gaps/not gaps'' along with the auxiliary language-model-based objective, such as minimizing the language model (LM) error when predicting the right answer for each gap.  Chinkina and Meurers~\cite{chinkina-meurers-2017-question} describe an automatic generation of exercises with a focus on particular linguistic forms. \comment{For instance, in a form exposure exercise, the target form is either a part of a question or is expected to be present in the learner's answer.  These exercises are implemented using part-of-speech tagging, parsing, resolving coreferences, and transformation rules for forming a question.}

\textbf{Distractors}, which are alternative, incorrect answers for multiple-choice exercises, can be generated using WordNets \cite{brown2005automatic,knoop2013wordgap,mitkov2003computer},\comment{lin2007automatic} ontologies \cite{papasalouros2008automatic,stasaski2017multiple}, N-gram co-occurrence \cite{hill2016automatic},\comment{ morphological generators~\cite{aldabe2006arikiturri},} and various similarity measures, e.g., morphological similarity \cite{pino2009semi} or distributional similarity from context, like word embedding similarity \cite{guo2016questimator,jiang2017distractor,kumar2015revup,susanti2018automatic,zesch2014automatic}.
Earlier methods (1) produced distractors for a target word based solely on its part of speech and similar frequency~\cite{brown2005automatic,shei2001followyou} or (2) used rules reflecting syntactic similarities between targets and distractors or explicit grammatical features that distractors should exhibit~\cite{chen2006fast,perez-beltrachini-etal-2012-generating}.  More realistic exercises can be generated by leveraging learner corpora to identify frequent errors as distractors.  For instance, Sakaguchi et al.~\cite{sakaguchi2013discriminative} propose a method of generating distractors using discriminative models trained on error patterns extracted from the English as a Second Language (ESL) corpus.  Panda et al.~\cite{panda-etal-2022-automatic} explores the use of round-trip neural machine translation to generate distractors for fill-in-the-blank exercises. By translating sentences to another language and back, the system produces a diverse set of challenging distractors, enhancing the quality of the exercises.

Several works focus on \textbf{distractor filtering and ranking}, to select the most suitable options from a set of generated candidates.  Filtering is done by feature-based ensemble learning (e.g., with random forest)~\cite{liang-etal-2018-distractor,murugan2022automatic}, neural rank models~\cite{murugan2022automatic,yeung2019difficulty}, or a combination of simple features with representations from pre-trained models, like BERT~\cite{gao2020distractor}.  Transformers have demonstrated good performance in ranking distractors by difficulty~\cite{yeung2019difficulty}.  Bitew et al.~\cite{bitew2022learning} showed that multilingual Transformer-based models for distractor retrieval and scoring can achieve superior results in selecting distractors compared to feature-based models.\comment{, as was demonstrated by manual evaluation.}

Wang et al.~\cite{wang2024automated} evaluates a method for automatically generating multiple-choice cloze questions using \textbf{large language models} (LLMs). The VocaTT engine, implemented in Python, processes target word lists, generates sentences and candidate word options using GPT-3.5 model, and selects suitable options.  In testing, 60 questions targeting academic words were generated, with expert reviewers judging 75\% of the sentences and 66.85\% of the word options as well-formed. 

Shen et al.~\cite{shen-etal-2024-personalized-cloze} introduces the Personalized Cloze Test Generation (PCGL) Framework, which utilizes LLMs to generate multiple-choice cloze tests tailored to individual proficiency levels.  Unlike traditional methods that require both a question stem and an answer, the PCGL Framework simplifies test creation by generating both the question stem and distractors from a single input word. The framework adjusts the difficulty level of the generated questions based on the learner's proficiency.

Several studies propose methodologies for exercise generation that are implemented in {\em actual language learning systems and applications}; here, we highlight a few examples.  The \textit{REAP} intelligent tutoring system~\cite{heilman2006classroom}, which can work with texts from any web source, incorporates cloze questions with and without distractors in its practice and assessment module~\cite{pino2008selection}.  \textit{FollowYou!} by Shei et al.~\cite{shei2001followyou} automatically generates language lessons from {\em authentic} texts uploaded by the learner.  This exercise generation system supports only the lexical level.  
\comment{Initially, the system assesses the learner's proficiency level through a diagnostic test and then generates relevant vocabulary exercises. Each time a new text is imported, the system creates a new set of exercises corresponding to the learner's current proficiency.}
The \textit{Lärka} learning platform~\cite{volodina-etal-2014-flexible} offers vocabulary and inflectional grammar exercises in Swedish based on single sentences extracted from a corpus. Exercises are tailored to the learner's specified proficiency level. 
\textit{WordGap}~\cite{knoop2013wordgap} is another method that creates cloze exercises from any given text or website.  Perez et al.~\cite{perez-cuadros-2017-multilingual} introduced a web-based \textit{Language Exercise App} to design multiple-choice, fill-in-the-gap, and shuffled sentence\footnote{Exercise which consists of ordering a list of shuffled words into a complete sentence.} exercises for Basque, Spanish, English, and French.  Heck and Meurers~\cite{heck-meurers-2022-parametrizable} present a grammar exercise generation feature within the \textit{FLAIR} search engine, which is capable of ranking texts relevant to studied linguistic constructs.  Users of the system can customize settings to define preferred exercise topics, e.g., ``Past tense'', and exercise types, e.g., fill-in-the-blank or drag-and-drop. \comment{ Depending on settings, exercises, distractors, and hints are automatically generated based on lemmatization, morphological generators, and specific rules.}

As was noted by Heck and Meurers~\cite{heck-meurers-2022-parametrizable}, incorporation of \textit{authentic} texts into exercise generation varies significantly across systems.  For example, while rule-based systems like \textit{Mgbeg}~\cite{almeida2017exercise} and \textit{GramEx}~\cite{perez-beltrachini-etal-2012-generating} do not employ authentic texts, others like \textit{FAST}~\cite{chen2006fast} and \textit{Lärka} utilize only individual sentences from authentic sources.  \textit{Revita}, \textit{Language Exercise App}, and \textit{FLAIR} integrate exercises seamlessly within authentic texts.

\paragraph{Reading} Various {\em reading comprehension} tasks, such as answering factual questions based on a text or determining the truth value of a statement, can be an integral component of a CALL system. Experimental results show that computer-generated questions can enhance learners' understanding of text~\cite{steuer2022educational}. Heilman~\cite{heilman2011automatic} addresses the generation of factual questions from any text, suitable for beginner and intermediate students.  The generated questions do not require domain-specific knowledge and solely test the ability to understand and retain information about the text.  Before neural networks, question generation was approached by performing lexical and syntactic transformations~\cite{mitkov2006computer}, using template-based methods~\cite{curto2012question}, or generating a question by replacing a target with a gap~\cite{becker-etal-2012-mind}.  Chinking and Meurers~\cite{chinkina-meurers-2017-question} focus on generating questions targeting linguistic forms and grammatical categories in text , in particular, form exposure questions (understanding of the target form is required to answer the question) and grammar concept questions (interpretation of a target form).  Exercises are generated using transformation rules, templates, and by replacing target forms with gaps.

A number of approaches address the task of automatic question generation using neural models, e.g., attention-based models~\cite{du-etal-2017-learning}, pointer networks~\cite{kumar2018automating,wang2018qg}, or Transformers~\cite{liu2020asking}.  Generating longer, semantically rich questions from passages, which are closer to real reading comprehension examination, was investigated in experiments by Gao et al.~\cite{gao2019generating} and Stasaski et al.~\cite{stasaski-etal-2021-automatically}. \comment{Another type of reading comprehension exercise is a true/false exercise, which allows evaluation of general understanding of learning material. Zou et al.~\cite{zou-etal-2022-automatic} propose an unsupervised True/False Question Generation approach, leveraging both template-based techniques and generative models to craft high-quality questions from a given passage. } Research on reading comprehension overlaps significantly with a vast field of Question Generation (QG) and Question Answering (QA).  Therefore, we leave related work on QG and QA outside the scope of this work. 

Recent advances in large language models (LLMs), such as GPT-3~\cite{brown2020language}, GPT-4~\cite{openai-report}, Llama~\cite{touvron2023llama}, Gemini~\cite{team2023gemini}, Mixtral~\cite{jiang2024mixtral}, etc. have demonstrated strong performance across various NLP tasks in zero-shot or few-shot settings.  For example, ChatGPT and similar dialogue models have significant potential to improve the generation of reading comprehension tasks.  Models can generate comprehension and expansion questions, provide summaries, provide explanations and definitions of unknown words, and adjust responses to suit learners' proficiency levels~\cite{kohnke2023chatgpt}.  

The study by Xiao et al.~\cite{xiao-etal-2023-evaluating} implemented a reading comprehension exercise generation system aimed at providing high-quality and personalized reading materials for middle-school English learners in China. The system used LLMs to generate reading passages and corresponding questions. Extensive evaluations, both automatic and manual, demonstrated that the system-generated materials were suitable for students and, in some cases, surpassed the quality of existing human-written content.

Säuberli and Clematide~\cite{sauberli-clematide-2024-automatic} explored how LLMs can be employed to generate and evaluate multiple-choice reading comprehension items. The study addressed the challenges of manual test creation and quality assurance by leveraging Llama 2 and GPT-4 to automate the process. The findings suggest that LLMs are capable of generating items of acceptable quality in a zero-shot setting. It was shown usage of LLMs is a promising approach for generating and evaluating reading comprehension test items, in particular for languages without large amount of data.

Nevertheless, concerns remain regarding the factual accuracy of generated responses, limited personalization, biases in the training data, potential plagiarism, and other ethical issues~\cite{caines2023application,kasneci2023chatgpt,nye2023generative,xiao-etal-2023-evaluating}.  As LLM technology evolves, many of these problems will be addressed, leading to an increased influence of LLMs in the educational sector.  For instance, a workshop Generative AI for Education (GAIED) at NeurIPS\footnote{https://gaied.org/neurips2023} is dedicated to the usage of LLMs for educational purposes, reflecting a commitment to address these concerns~\cite{denny2024generative}.  

\paragraph{Listening and speaking} {\em Listening skills} can be improved through exposure to authentic native audio materials.  The use of fully or partially synchronized captions of audio and video content can help the learner improve comprehension, as well as promote the acquisition of vocabulary~\cite{mirzaei2017partial,montero2014effects,perez2014less}.  While many AI-powered applications like YouTube, Spotify, Siri, Alexa, etc. offer opportunities to improve listening comprehension, they are not developed for learning.  Dizon~\cite{dizon2020evaluating} experimentally showed that using Alexa did not lead to improvements in listening comprehension.  Several works~\cite{vu2022integrating,julyati2023use} report on experiments, which were run with two groups of students: one group using a CALL application and a control group.  All experimental groups showed improvement in listening comprehension skills.  Lebedeva et al.~\cite{lebedeva2017computer} showed that using CALL applications significantly improves selective listening\footnote{Focusing attention only on some specific information.} skills.  It could potentially be explained by the fact that individual work with an educational platform allows teachers to better understand individual learning needs and challenges.

Listening comprehension exercises can be generated in a manner similar to reading comprehension.  For example, exercises by \textit{Duolingo}~\cite{nushi2017duolingo}, where students should type what they hear, are based on matching the students' input against the original text. \textit{PushkinOnline} presented by Lebedeva et al.~\cite{lebedeva2017computer} includes post-listening tasks that focus on factual information in the audio text.  The \textit{Listening Hacked} application~\cite{vu2022integrating} facilitates learning by watching movies and intermittently transcribing selected utterances.  \comment{This application also annotates all transcribed segments and, in the case of errors, it automatically suggests listening to target words in various contexts.  If the student skips some utterances, the application incorporates exercises based on these skipped segments into a page for personalized practice.}

An et al.~\cite{an2024funaudiollm} introduce FunAudioLLM, a framework designed to enhance natural voice interactions between humans and LLMs. It integrates two models: SenseVoice for multilingual speech recognition and emotion detection, and CosyVoice for natural speech generation. The system enables applications like speech-to-speech translation, interactive podcasts, and expressive audiobook narration, enhancing listening practice through expressive and context-aware audio content. It is not developed for language learning, but can have wide application for improving listening comprehension. 

Another emerging area is \textit{dialogue systems} for educational purposes.  Dialogue systems equipped with Automatic Speech Recognition (ASR) and Generation (ASG) have the potential to meet the need for engaging, goal-oriented, always-available interaction.  Such systems offer an environment, where students can make mistakes without fear of being judged by an interlocutor~\cite{cucchiarini2018second,golonka2014technologies}.  In a conversation with a dialogue system, the learner can repeatedly practice the same content and train for quick reactions to audio input.\comment{which is a critical skill often evaluated at exams.}  Various studies have demonstrated that interacting with dialogue systems can reduce speaking anxiety~\cite{bashori2022web,chen2022effects,hapsari2022ai} and enhance receptive vocabulary~\cite{bashori2022look}.  

Sharma et al.~\cite{sharma2024comuniqa} propose Comuniqa, an LLM-based system designed to enhance English speaking skills, particularly in non-native contexts. The study evaluates the system's effectiveness compared to human tutors, finding that while LLMs offer scalable speaking practice opportunities, they currently lack the nuanced feedback and empathy provided by human instructors. 

Xu et al.~\cite{xu2024large} present situational dialogue models fine-tuned based on LLMs to facilitate conversational practice in various scenarios. The models aim to combine open-ended conversation with focused, context-specific tasks, providing learners with opportunities to practice speaking about diverse topics. 

An extensive review of research on speech-recognition chatbots for language learning is presented by Jeon et al.~\cite{jeon2024systematic}.

\subsection{Assessment and feedback}

Assessment of learner input is closely tied to the Student and Instruction models.  The Student model can be updated based on the history of correct and incorrect answers.  Using this data, the Instruction model determines the next steps in topic or skill development and provides tailored feedback to the learner.

In evaluating learner input, various methods can be applied, such as pattern matching, rule-based methods, and statistical methods.  Pattern matching is particularly suitable for assessing answers in multiple-choice exercises, which typically have only one correct answer.  However, exercises like fill-in-the-gap and sentence shuffling might have multiple valid answers.  Therefore, models that are capable of assessing grammatical correctness should be incorporated into a CALL system.

Grammatical error detection (GED) plays an important role in evaluating written (or spoken) text, and impacts scoring and the provision of feedback on grammatical tasks.  Accurate detection of grammatical errors is crucial\comment{not only for assigning correct scores but also} for assessing the learner's understanding of specific grammatical constructs.  This, in turn, influences the Instruction model and the choice of exercises for future practice. 

Automatic Essay Scoring (AES), or evaluation, refers to the assessment of written essays by the CALL system.  It is a wide research area, dating back to the 1960's~\cite{ramesh2022automated,vijaya2022essay}.\comment{ke2019automated}  AES is a highly challenging task as it requires assessing grammar and spelling together with semantics, discourse, and pragmatics.  Research on AES focuses on providing a {\em holistic score} that reflects the overall quality of the learner's text, and on providing {\em dimensional scoring}---analytic scores assigned to different dimensions of evaluation~\cite{carlile2018give,mathias2020can,phandi-etal-2015-flexible,somasundaran2014lexical}.\comment{zhang2019co}  Research on AES examines style, i.e., spelling errors, length, etc.~\cite{mathias-bhattacharyya-2018-thank,phandi-etal-2015-flexible},\comment{lu2017developing} content and coherence, i.e., meaning and similarity with graded essays~\cite{chen2018relevance,dong-zhang-2016-automatic,taghipour-ng-2016-neural},\comment{ratna2019term,amalia2019automated} and multiple attributes at once~\cite{jin-etal-2018-tdnn}.\comment{alghamdi2014hybrid,al2017automated}  Further details on research in the area of AES can be found in the work of Lim et al.~\cite{lim2021comprehensive}.

Although some AES systems that have achieved success are freely available, such as the Automated Writing Placement System~\cite{yannakoudakis2018developing} integrated into Cambridge English Write \& Improve\textsuperscript{TM},\footnote{https://writeandimprove.com} most systems are proprietary, e.g., \textit{Project Essay Grader} (PEG)~\cite{page1966imminence}, \textit{Intelligent Essay Assessor} (IEA)~\cite{foltz1999intelligent},\comment{pearson2010intelligent} \textit{IntelliMetric}\footnote{https://intellimetric.com/direct/}, and \textit{E-rater}~\cite{attali2006automated}.  Among other systems, for instance, Zupanc et al.~\cite{zupanc2017automated} propose the \textit{SAGE} system, which \comment{assesses the coherence and consistency of learner text.  \textit{SAGE}} evaluates coherence by analyzing vocabulary variation in the text.  \comment{Consistency is evaluated based on semantic errors, with the help of information extraction and logical reasoning tools.  \textit{SAGE} provides feedback by highlighting inconsistent text sections.}  \textit{Revision Assistant}~\cite{woods2017formative}, a data-driven educational tool, employs an essay scoring model that detects sentences that require improvement.\comment{and assigns a score to various text traits, such as the use of evidence, clarity, and organization. } 
Ramesh et al.~\cite{ramesh2022automated} presented a review of other academic and commercial ARS systems based on evaluation techniques, feature extraction methods, accuracy, and comparison with other systems.  Despite these advancements, to the best of our knowledge, no AES system has been integrated into an existing CALL system.

AES is typically approached as a supervised machine learning problem and requires a training set of human-graded essays.  Initially, the development of AES focused on handcrafted discrete features~\cite{chen2013automated,dikli2006overview,phandi-etal-2015-flexible,vajjala2018automated,wang2008automated}.\comment{, with some studies exploring the most predictive linguistic feature~\cite{vajjala2018automated}. The field has since evolved to automatically extract syntactic and semantic features using machine learning techniques.}  Various works have proposed regression- or classification-based models that employ features such as character, word, sentence counts, parts of speech, n-grams, misspelled words, and syntactic features, \comment{like the depth of a parse tree} etc.~\cite{cummins-etal-2016-constrained,persing-ng-2013-modeling,sakaguchi-etal-2015-effective,salim2019automated}.  Scoring essays by measuring the similarity of student responses to reference responses was proposed by Sultan et al.~\cite{sultan-etal-2016-fast} and S{\"u}zen et al.~\cite{suzen2020automatic}. To evaluate essay semantics, techniques like Latent Semantic Analysis~\cite{islam2010automated,sendra2016enhanced},\comment{shermis2013contrasting,al2019automated} Latent Dirichlet Allocation~\cite{kakkonen2006applying}, or neural latent semantic model~\cite{tashu2022deep} have been applied.
    
Recent methods are based on deep-learning models, such as recurrent and convolutional neural networks~\cite{alikaniotis-etal-2016-automatic,dong-zhang-2016-automatic,dong2017attention,taghipour-ng-2016-neural,zhang-litman-2018-co}, combinations of kernel functions and word embeddings~\cite{cozma-etal-2018-automated}, and Transformers~\cite{fiacco-etal-2023-towards,lun2020multiple,mayfield2020should,sethi2022natural}.\comment{rodriguez2019language}  Some effective solutions combine multiple linguistic features with neural models~\cite{dasgupta-etal-2018-augmenting,liu2019automated,tay2018skipflow}.  Neural essay scoring systems presented by Cummins et al.~\cite{cummins2018neural} and Nadeem et al.~\cite{nadeem2019automated} employ a multitasking framework, with GED as an auxiliary task.  \comment{Uto et al.~\cite{uto2020robust} introduced a framework that integrates a deep neural network with Item Response Theory models.}

Liew and Tan~\cite{10.1145/3709026.3709030} explore the potential of various LLMs, including GPT-4, GPT-3.5, PaLM, and LLaMA2, in automated essay scoring tasks. The research evaluates the performance of these models in assessing essays, highlighting their capabilities and limitations in educational settings. The study shows that LLMs can achieve substantial agreement with human markers in AES and highlights LLMs potential to enhance educational assessment practices.

Cai et al.~\cite{cai2025rank} introduces the "Rank-Then-Score" (RTS) framework, a novel fine-tuning method for LLMs aimed at improving essay scoring capabilities. RTS combines two specialized LLMs: one fine-tuned for essay ranking and another fine-tuned for scoring. The approach is evaluated on both Chinese and English datasets, showing enhanced performance over traditional methods. 

\comment{The performance of AES systems is evaluated based on the comparison of the system's scores vs.~experts' scores.  AES systems improve over time, offering significant reductions in the labor-intensive manual essay grading.  Additionally, these systems ensure consistent evaluation criteria.  Moving forward from manually crafting features and developing more sophisticated language models facilitates the creation of systems, which are tailored to specific domains, genres, prompts, etc.  Nevertheless, challenges persist in AES, such as the system's inability to assess the writer's creativity or the possibility of "gaming" the system by students~\cite{hussein2019automated}.  Another concern relates to automatically extracted features: some of the evaluated features might not be related to parameters that should be assessed, e.g., personal style characteristics.}

{\em Feedback} provided by AES systems is crucial, because it must be {\em formative}, interpretable, and accurate to effectively aid learners in improving their writing skills.  Multiple approaches concentrate on evaluating the quality of automatically generated feedback\comment{, benchmarking it against feedback from teachers}~\cite{liu2016automated,woods2017formative}.\comment{dikli2010nature}  Studies show that while automatic feedback can enhance the quality of writing, it suffers from inferior quality compared to teacher feedback and tendency to be too general or to include too many suggestions.  On the other hand, automated feedback is immediate and can stimulate more attempts to rewrite the text.  \comment{Some research also focuses on generating feedback that is not only high-quality but also fluent in the form of utterances; they approach it by leveraging deep generative models~\cite{lu2021integrating}.}

\paragraph{Automatic speech assessment}

Enhancing speaking skills is one of the key components of the teaching curriculum, designed to improve learners' ability to communicate effectively in the target language.  However, finding opportunities to practice speaking outside the classroom can be challenging.  Additionally, teachers often have limited time to interact individually with each student.  CALL systems equipped with Automatic Speech Recognition offer an opportunity for potentially unlimited  L2 speaking practice, together with automatic feedback on various aspects of speech.

Bibauw et al.~\cite{bibauw2019discussing} categorized dialogue-based CALL systems into several types.  They include intelligent tutoring systems, which assess the learner's speech and provide feedback; computer-assisted pronunciation training (CAPT) systems for enhancing pronunciation accuracy; spoken dialogue systems and conversational
agents (SDS/CA); and chatbots.  SDSs focus on dialogue management and multimodality, typically confined to goal-oriented conversations.  Chatbots are typically text-based systems designed to generate reactive responses~\cite{hsu2021proposing}.  

Automatic speech assessment is a vast research area, part of which concentrates on the automated assessment of non-native speaking proficiency.  This area of study\comment{, which began over 30 years ago,} initially centered on assessing pronunciation quality by comparing the learner's scripted responses to native speech models~\cite{bernstein1990automatic,witt2000phone}.  Subsequent research progressed towards evaluation of {\em spontaneous} speech, emphasizing various speech attributes, e.g., grammatical errors in spontaneous speech~\cite{knill2019automatic,lu2019impact}, pronunciation quality~\cite{yoon2019content}, lexical errors and accent~\cite{kyriakopoulos2020automatic}, or off-topic responses~\cite{lee2017off,raina2020complementary,wang2019automatic}.

As an example of a system that provides detailed feedback on learner speech, we can consider \textit{SpeechRater}\textsuperscript{SM}~\cite{chen2018automated,zechner2009automatic}, an automated scoring service, operational since 2006.  It stands out as the first system to score open-ended, spontaneous non-native English speech.  It is based on ASR models and NLP technology to filter off-topic or non-English responses and to provide feedback on prosody, pronunciation, vocabulary, and grammar.  \comment{Together, these feedback dimensions represent the construct coverage of the TOEFL iBT speaking test, which includes delivery, language use, and topic development \cite{bridgeman-2012-toefl}.}

Fu et al.~\cite{fu2024pronunciation} introduce a scoring system based on LLMs for evaluating learners' speech across various dimensions such as accuracy and fluency. The system maps the learner's speech into contextual features, which are then aligned with text embeddings to predict assessment scores. Experiments demonstrate that the proposed system achieves competitive results compared to existing baselines on the Speechocean762 dataset.  

Wang et al.~\cite{wang2023assessing} focus on evaluating phrase breaks in ESL learners' speech using both pre-trained language models and LLMs. The study introduces methods for overall assessment of phrase breaks and fine-grained evaluation at each possible break position, demonstrating the potential of LLMs in capturing prosodic features of speech. 

The challenges associated with generating immediate and personalized feedback in automatic speech assessment are similar to those in essay scoring~\cite{yoon2019toward}. Several factors influence whether automatic feedback is useful for learners~\cite{ranalli2018automated}.  These factors include {\em explicitness}---whether the feedback is direct or indirect.  Indirect feedback merely indicates the presence of an error, without detailing its nature or correcting strategies; it is generally less effective than explicit direct feedback~\cite{ellis2006implicit}.  The {\em accuracy} of feedback significantly impacts the student's perception of the usefulness of the entire learning tool.  Additionally, most existing tools do not account for {\em individual differences} among users for feedback generation, though L2 proficiency and educational background can significantly affect how learners interpret feedback~\cite{hyland200611,oneill2019stop}.

\section{Conclusions and Future Work}
\label{conclusions}

We present a survey of the AI methods which can contribute to language learning, viewed from the point of view of a developer of a CALL system.  We highlight the interconnection between the structural components of ITSs and their connection with system functionality, such as automatic generation of exercises, assessment, and providing feedback.  We review in depth\comment{???} the methods for modeling learner mastery, the most common approaches to automatic generation of exercises and assessment in relation to various language skills.

The survey is dense and eclectic by design and by necessity---a great deal of research is still required to arrive at coherent and comprehensive theories of intelligent support for language learning.  Such theories are still lacking, and surveys are one step toward developing such theories in the future.

In future work, we plan to address further facets, such as the level of learner proficiency, and contrast the approaches for beginners vs.~advanced learners.  We plan to cover the existing approaches to {\em automatic readability assessment}\comment{???}---estimating the difficulty of a text,---exercises, and other learning content, as well as research in the area of text simplification.  An important area to be reviewed in the future is the {\em creation} of the resources necessary for language learning, e.g., annotations, and other data that might be produced by learners (or teachers) while interacting with ITS, implicit crowd-sourcing of learner corpora, and other crowd-sourcing approaches.  We plan to analyze existing commercial and academic ITS/CALL systems using the facets presented in this paper.





\bibliography{thesis_refs}


\end{document}